\documentclass[conference]{IEEEtran}
\IEEEoverridecommandlockouts
\usepackage{microtype}
\usepackage{graphicx}
\usepackage{subfigure}
\usepackage{booktabs} 
\usepackage{mathtools}
\usepackage{comment}

\usepackage{newunicodechar}

\usepackage{algorithm}
\usepackage[table,xcdraw]{xcolor}
\usepackage{multirow}
\usepackage{cite}
\usepackage{amsmath,amssymb,amsfonts}
\usepackage{algorithmic}
\usepackage{textcomp}
\usepackage{xcolor}
\usepackage{xspace}
\usepackage{fancyhdr}
\usepackage[hyphens]{url}
\usepackage{hyperref}
\usepackage{caption}
\usepackage{ragged2e}
\usepackage{makecell}

\usepackage{alphabeta}
\usepackage[utf8]{inputenc}

\usepackage{tikz}
\usepackage{cite}
\usepackage{amsmath,amssymb,amsfonts}
\usepackage{algorithmic}
\usepackage{graphicx}
\usepackage{textcomp}
\usepackage{xcolor}
\usepackage[a4paper, total={184mm,239mm}]{geometry}
\newcommand{\PCignore}[1]{}

\def\Snospace~{\S{}}

\newcommand{\squishlist}{
 \begin{list}{$\bullet$}
  { \setlength{\itemsep}{0pt}
     \setlength{\parsep}{3pt}
     \setlength{\topsep}{3pt}
     \setlength{\partopsep}{0pt}
     \setlength{\leftmargin}{1.5em}
     \setlength{\labelwidth}{1em}
     \setlength{\labelsep}{0.5em} } }

\newcommand{\squishlisttwo}{
 \begin{list}{$\bullet$}
  { \setlength{\itemsep}{0pt}
     \setlength{\parsep}{0pt}
    \setlength{\topsep}{0pt}
    \setlength{\partopsep}{0pt}
    \setlength{\leftmargin}{2em}
    \setlength{\labelwidth}{1.5em}
    \setlength{\labelsep}{0.5em} } }

\newcommand{\squishend}{
  \end{list}  }

\newcommand\circled[1]{\tikz[baseline=(char.base)]{
            \node[shape=circle,fill=darkgray,inner sep=0.5pt] (char) {\textcolor{white}{#1}};}}

\newcommand{\system}{{\sf{DiGamma}}\xspace}
\newcommand{\coopt}{{\sf{Co-opt Framework}}\xspace}

\def\BibTeX{{\rm B\kern-.05em{\sc i\kern-.025em b}\kern-.08em
    T\kern-.1667em\lower.7ex\hbox{E}\kern-.125emX}}
\begin{document}

\title{DiGamma: Domain-aware Genetic Algorithm for HW-Mapping Co-optimization for DNN Accelerators\vspace{-0.3cm}\thanks{This work was supported by NSF Award 1909900. This work is available at https://github.com/maestro-project/digamma.}}

\author{
\IEEEauthorblockN{Sheng-Chun Kao\IEEEauthorrefmark{1}, Michael Pellauer\IEEEauthorrefmark{2}, Angshuman Parashar\IEEEauthorrefmark{2}, Tushar Krishna\IEEEauthorrefmark{1}} \\
\vspace{-4mm}
\IEEEauthorblockA{\IEEEauthorrefmark{1} Georgia Institute of Technology \IEEEauthorrefmark{2} NVIDIA \vspace{-0.4cm}} \\
\IEEEauthorblockA{\IEEEauthorrefmark{1} \emph{skao6@gatech.edu, tushar@ece.gatech.edu} 
\IEEEauthorrefmark{2} \emph{ \{mpellauer, aparashar\} @nvidia.com}}
\vspace{-10mm}
}

\maketitle
\begin{abstract}
The design of DNN accelerators includes two key parts: HW resource configuration and mapping strategy. Intensive research has been conducted to optimize each of them independently.
Unfortunately, optimizing for both together is extremely challenging due to the extremely large cross-coupled search space.
To address this, in this paper, 
we propose a HW-Mapping co-optimization framework, an efficient encoding of the immense design space constructed by HW and Mapping, and a domain-aware genetic algorithm, named \system, with specialized operators for 
improving search efficiency. 
We evaluate \system with seven popular DNNs models with different properties. Our evaluations show \system can achieve (geomean) \textbf{3.0x} and {10.0x} speedup, comparing to the best-performing baseline optimization algorithms, in edge and cloud settings.
\end{abstract}
\vspace{-2mm}
\section{Introduction}
\captionsetup[figure]{font=small}
\captionsetup[table]{font=small}
\vspace{-1mm}

Specialized DNN accelerators design has become a hot topic and kept breaking through state-of-the-art performance for both training and inference~\cite{eyeriss_isca, nvdla, du2015shidiannao}. The efficiency of a DNN accelerator is decided by its hardware (HW) resource configuration and the applied DNN mapping (dataflow + tiling) strategy, where both have been shown independently to be able to impact the accelerator performance by several orders~\cite{confx, gamma, maestro}. 

\begin{figure}[t]
\begin{center}

\includegraphics[width=1\linewidth]{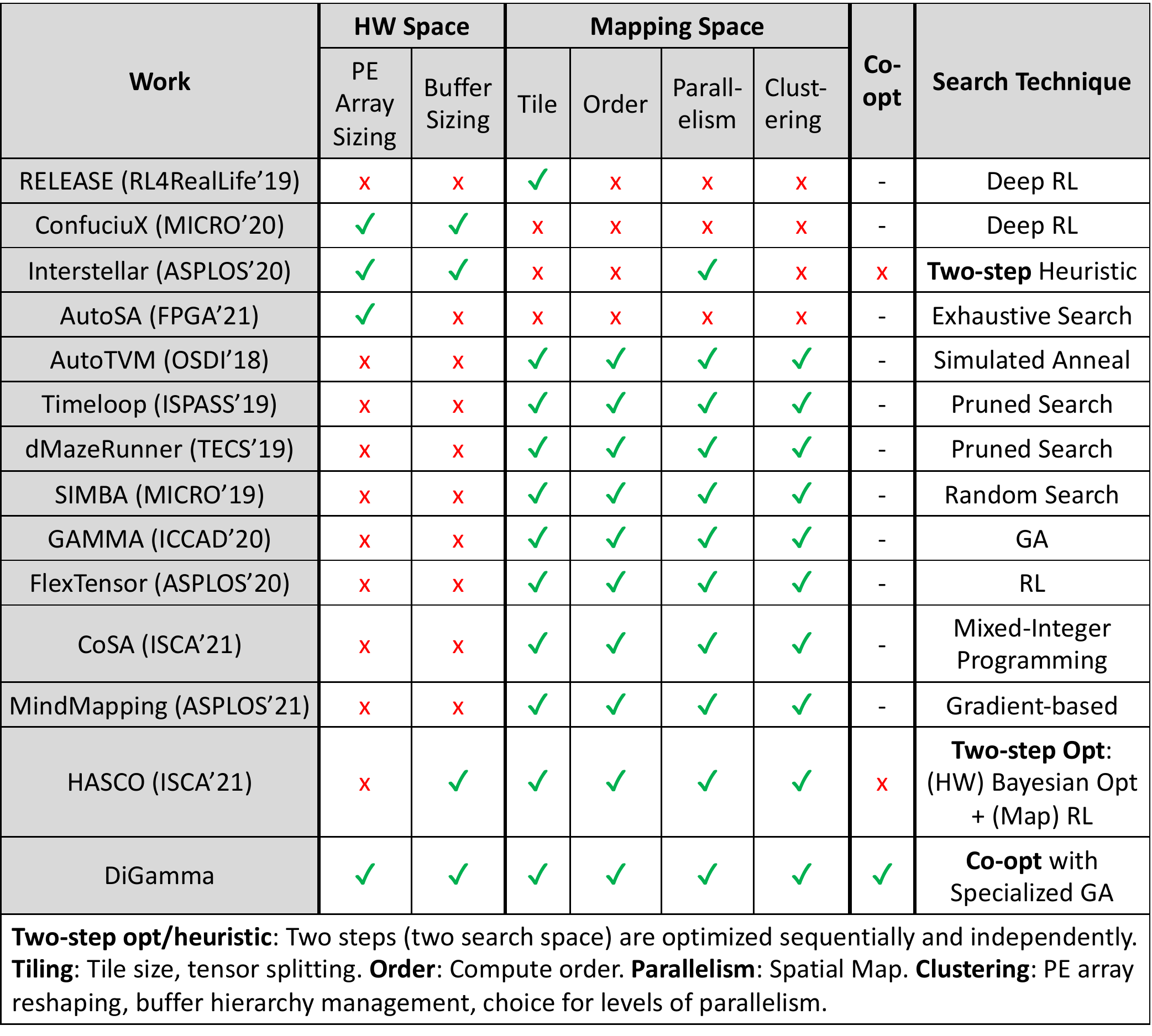}
\end{center}
\vspace{-0.3cm}
\caption{State-of-the-Art HW and Mapping optimization frameworks. }

\label{fig:related_work_table}
\end{figure}

The design space for HW configuration and mappings is extremely large. 
For a given accelerator with fixed HW resources, the number of ways to map the DNN computation on it can be as large as $O(10^{24})$~\cite{gamma}. Meanwhile, for a 
given mapping strategy and fixed chip area budget, there can be $O(10^{12})$ possible HW implementations~\cite{confx}.
We elaborate on these in \autoref{sec:challenge}. 
Many accelerators today either heavily rely on (i) heuristic and expert knowledge to strike the balance of resource allocation between compute resource, memory, and mapping~\cite{nvdla, eyeriss_isca, du2015shidiannao} (manual-tuned \textit{Fixed HW-Mapping}), or (ii) using AI/ML methods for finding efficient mappings given fixed hardware at compile-time~\cite{gamma} (\textit{Mapping-opt}), or 
(iii) sizing the hardware resources at design-time assuming a fixed mapping strategy~\cite{confx} (\textit{HW-opt}). We summarize the related works in \autoref{fig:related_work_table}. One main challenge of these schemes is that they still include full or part of human-in-the-loop manual-tuning process for deciding a fixed mapping, a fixed HW, or both, which stacks up engineering cost. 
On the flip-side, without putting in this costly repeated manual-tuning effort, DNN accelerators may end up sub-optimal performance once the workload (the target DNN model) changes.

With the growth of AutoML, optimizing HW and Mapping together automatically with AI/ML offers a potential solution.
However, how to effectively co-optimize the HW and mapping is still an open question.
%
%
%
To address this, we propose a HW-Mapping co-optimization framework (\coopt) and an optimization algorithm, named \system (\digamma), which search the HW resources configuration and the optimized mapping simultaneously. 
We make the following contributions:

\squishlist

\item We propose a HW-Mapping co-optimization framework (\coopt), which takes in any DNN model(s), design objective, budget, and constraint, and generates an accelerator design point, HW (i.e., numbers of PEs, number of memory levels, sizes of buffers at each level) and mapping (i.e., parallelism, loop order, clustering, tile sizes). We abstract the detail of performance modeling for different DNNs, chip constraints and provide a generic interface, where many existing optimization algorithms can be plugged in, as shown via our experiments in \autoref{sec:other_opt}. 

\item We propose an efficient design point encoding, which describes both HW and mapping with a list of parameters. Our encoding method constructs a compact representation of the cross-coupled design space that boosts the efficiency of the optimization algorithms.

\item We propose a domain-aware genetic algorithm-based optimization method, named \system (\digamma). It is specifically designed for HW-Mapping design space and comes with specialized optimization operators to step through the design space in a structured manner, and its HW exploration strategy respects the interaction between HW and mapping.

\squishend


\vspace{-1mm}
We provide evaluations across two different settings: edge and cloud platform resources, and seven popular DNN models with diverse characteristics. The evaluation shows \system can achieve (geomean) \textbf{3.0x} and \textbf{10.0x} speedup, comparing to the best-performing baseline optimization algorithms, in edge and cloud settings, respectively, and (geomean) \textbf{1.25x} and \textbf{2.0x} speedup over the best-performing baseline HW optimization or Mapping optimization scheme, in edge and cloud settings.

\vspace{-2mm}
\section{Background, Motivation and Related Work}

\subsection{DNN Accelerators}
DNN accelerator design points can often be described with two parts: HW resources and mapping strategy, described next.

\subsubsection{\textbf{Hardware Resources}}
Spatial DNN accelerators comprise an array of Processing Elements (PE). Each PE has a MAC to compute partial sums, and a local buffer (L1) to store weights, activations, and partial sums.
The accelerators also house a shared global buffer (L2) to pre-fetch activations and weights from off-chip memory for the next tile of computation that will be mapped over the PEs and L1, as shown later in \autoref{fig:system}(d). Networks-on-Chip (NoCs) are used to distribute operands from L2 to L1 inside each PE and collect the partial or full outputs and write them back to the L2.

\subsubsection{\textbf{Mapping}}
A \textit{mapping}~\cite{maestro} is comprised of: (1) tiling (how tensors are sliced, stored, and fetched across the memory hierarchy), (2) compute order (order in which loop computations are performed), (3) parallelism (how compute is mapped across PEs in space), and (4) clustering (how compute/buffer are structured into hierarchy of levels). In literature, (2)(3)(4) together are often called \textbf{\textit{dataflow}}~\cite{eyeriss_isca,maestro}. 

\vspace{-2mm}
\subsection{Prior work in DNN Accelerator Design Space Exploration}
\label{sec:related_work}

\textbf{HW Configuration Optimization.}
A typical HW resource optimizer~\cite{confx} works as follows. It takes in a target DNN, the mapping strategy, an optimizing objective (e.g., latency, power) and resource constraints (e.g., area budget) as input. It returns a HW resource allocation (PEs and buffer sizes). The HW resource allocation problem has been widely-studied at \textit{compile-time} in the FPGA community and 
at \textit{design-time} for DNN ASIC accelerators. Some recent works apply ML techniques to the HW allocation problem such as reinforcement learning (RL)~\cite{confx}, GNN~\cite{google_ai_chipdesign}, and others (\autoref{fig:related_work_table}).

\textbf{Mapping Optimizations.}
A typical mapping optimizer~\cite{gamma} takes in a target DNN, the \textit{ accelerator's HW configuration (i.e., resources)} as constraint, and an optimizing objective (e.g., latency, power). It can be used at \textit{compile-time} or even at \textit{run-time} (for reconfigurable accelerators). The mapping optimization process includes techniques such as formulating a more comprehensive design space for better capturing the accelerator behavior, improving the design space description for search efficiency, and incorporating different ML techniques in the optimization process~\cite{gamma, simba} (\autoref{fig:related_work_table}).

\vspace{-2mm}
\subsection{Challenges with HW-Mapping Co-optimization}
\label{sec:challenge}
A HW-Mapping co-optimizer can enable mapping-aware HW design and further optimize the accelerator design at \textit{design-time}. However, designing a HW-Mapping co-optimizer is not trivial. The design space is the cross-product of HW space (as large as $O(10^{12})$\footnote{Assuming PEs:128x128 and maximum buffers:100MB, the number of combinations is $O(10^{12})$.}) and mapping space (as large $O(10^{24})$~\cite{gamma}), which can lead to an design space as large as $O(10^{36})$. Therefore basic techniques like exhaustive searches become impractical. An optimization-based algorithm (e.g., RLs, GA, simulated annealing, and so on) is needed. A naive optimization-based HW-Mapping co-optimizer can be formulated as follows. One can formulate a two-loop optimization process, where the outer-loop optimizes the HW, and the inner-loop (takes in the HW parameters from outer-loop) optimizes the mapping or vice versa. For e.g., a highly tuned mapper GAMMA~\cite{gamma} requires about 10 mins to converge to a mapping solution of a given HW configuration. For a two-loop optimization, the found solution at inner-loop (mapper) becomes the feedback of one single sampling point of HW optimizer at outer-loop, where outer-loop can easily require more than 10K sampling points. A naive two-loop optimization requires 1.6M sampling points and more than 1,600 hours, which is challenging for practical usage.

\begin{figure}[t]
\begin{center}

\includegraphics[width=1\linewidth]{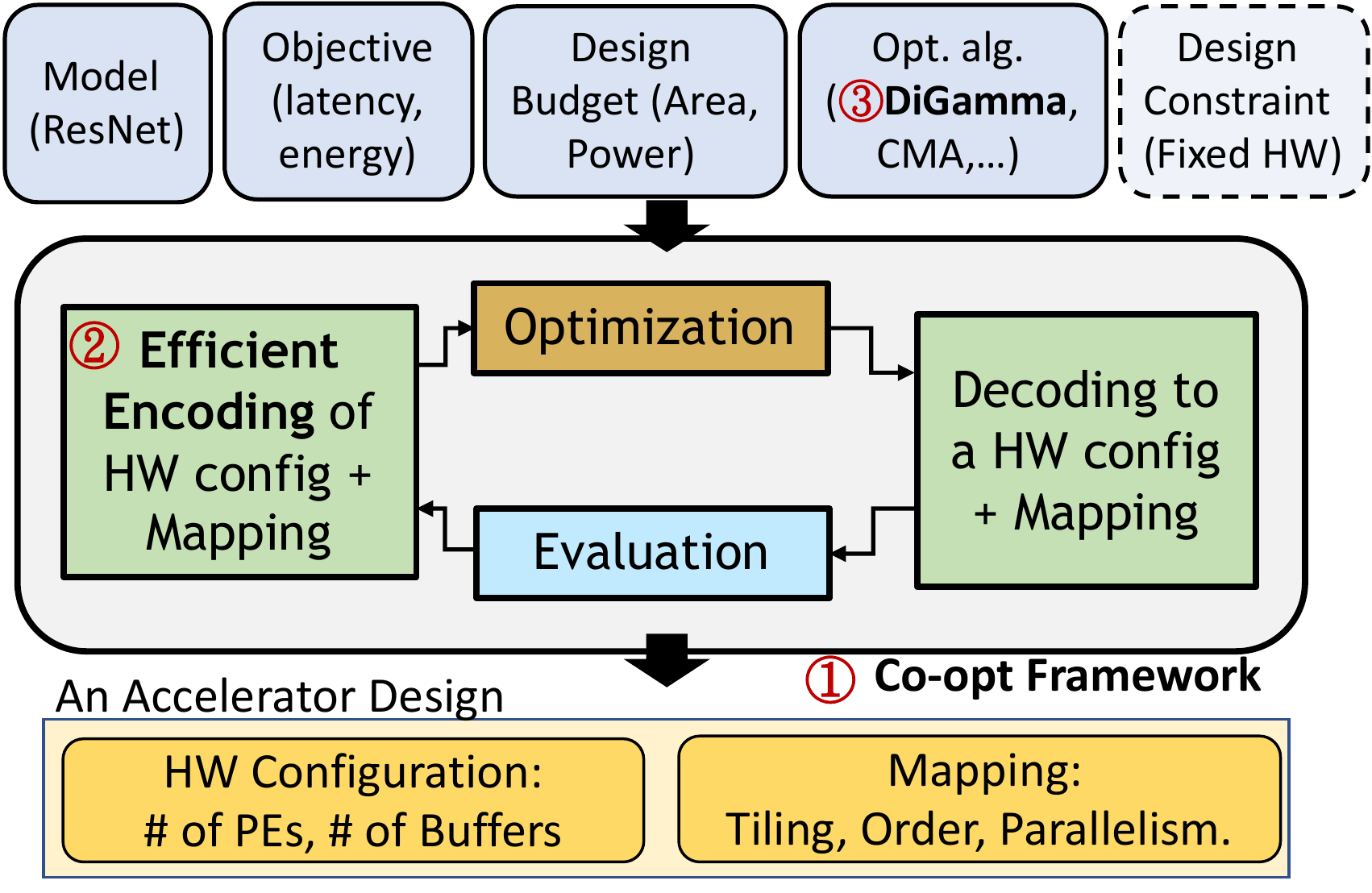}
\end{center}
\vspace{-0.3cm}
\caption{HW/Mapping Co-optimization Framework. Our three technical contributions are highlighted.}
\vspace{-0.1cm}
\label{fig:toolflow}
\end{figure}

\begin{figure*}[t]
\begin{center}

\includegraphics[width=1\linewidth]{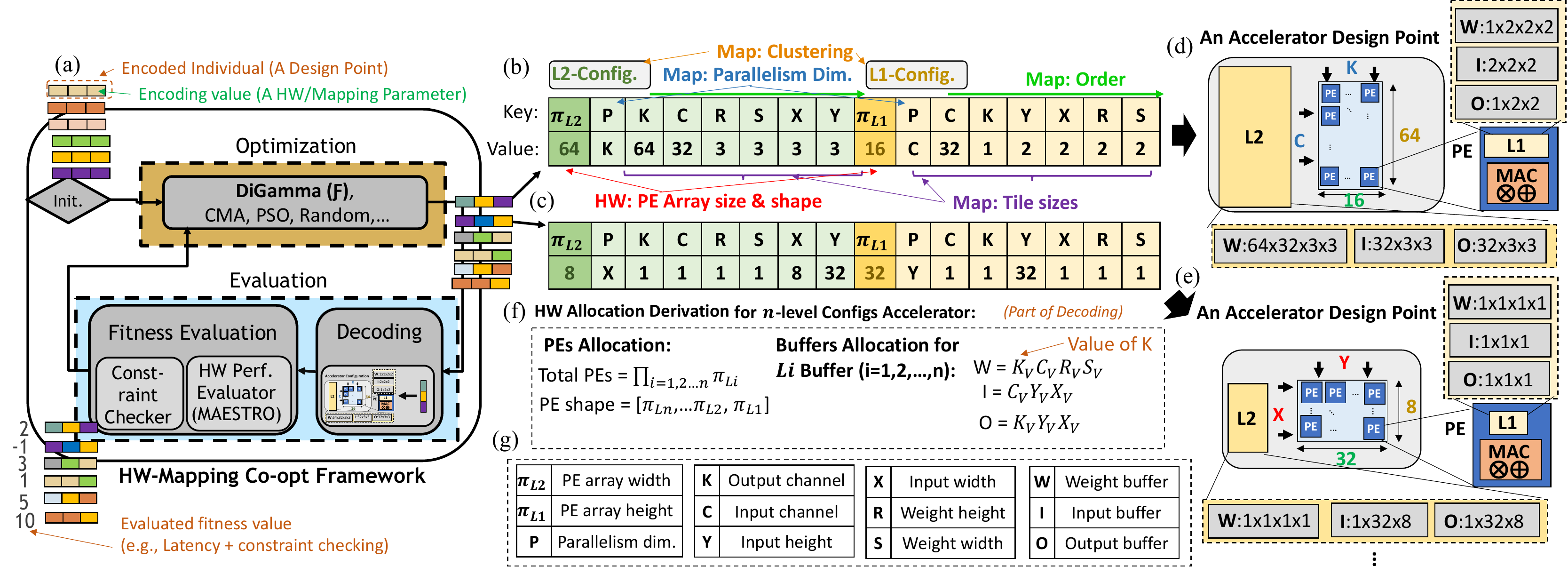}
\end{center}
\vspace{-0.4cm}
\caption{(a) \coopt, (b-c) the HW-Mapping encoding representation, and (d-e) the corresponding decoded accelerator configuration. (f) The formula for calculating minimum on-chip buffer requirement. (g) The definition of notations.}
\vspace{-0.7cm}
\label{fig:system}
\end{figure*}

\vspace{-2mm}
\section{Technical Approach}
\label{sec:methodology}

\textbf{Problem Formulation.}  \emph{Under a design budget (e.g., chip area) and given a DNN model, we aim to design an accelerator with optimized HW resource configuration for PEs, local buffer (L1), and global buffer (L2), and an optimized mapping strategy.}
\vspace{-1mm}

\vspace{-2mm}
\subsection{High-level Overview}
We integrate the two searching loops (\autoref{sec:challenge}), HW and mapping, into one unified optimization process, and propose \circled{1} a HW-Mapping co-optimization framework (\coopt), as shown in \autoref{fig:toolflow}. Note that there are two main factors deciding the effectiveness of an optimization framework: efficiency of the design-point encoding and efficiency of the optimization/ search algorithm. In this work, we propose \circled{2} an efficient encoding method for HW configuration and mapping (\autoref{sec:encoding}) and \circled{3} a sample-efficient optimization algorithm (\system) (\autoref{sec:algorithm}).

\vspace{-2mm}
\subsection{HW-Mapping Co-optimization Framework}
\label{sec:copopt}
\coopt takes the input of target model, optimization objective, design budget, an optimization algorithm, and (optionally) a design constraint, and generates an optimized accelerator design point with HW configuration and mapping strategy.
The design constraint is optional, for supporting two additional use-cases:  1) Fixed-HW: when the researcher/ engineer already has a designed accelerator and only wants to search for an optimal mapping; 2) Fixed-Mapping: when the researcher has a manual-tuned mapping (e.g., NVDLA~\cite{nvdla}) and wants to understand the optimal HW configurations for designing a specialized accelerator (e.g., understanding the compute to memory balance). \coopt can deal with these constraints by restricting the design space accordingly. 
As shown in \autoref{fig:system}(a), \coopt includes an optimization block, where different optimization algorithms can be plugged and played, and an evaluation block, where the proposed design points are evaluated and scored to guide the algorithms.

\subsubsection{\textbf{Optimization Block.}} 
One main goal of this work is to develop a generic framework for HW-Mapping co-optimization that is not tiled with any optimization/searching algorithms. We abstracted the underlying detail of taking different DNN models as inputs, understanding different design budgets, encoding/decoding of HW and mapping, and so on, and expose a generic interface for all the optimization algorithms. The only task left for any optimization algorithm (from any optimization library~\cite{nevergrad} or custom-designed) is to find a list of parameters (together forming a design point) that yields the highest reward (fitness). The sampling budget, which is the number of design points the algorithms are allowed to sample, is a hyper-parameter of optimization block that can be set by the users. It controls the number of optimization loops (optimization, evaluation, optimization,...) in the framework.

\subsubsection{\textbf{Evaluation Block.}}
The evaluation block includes a decoding module (described in \autoref{sec:decoding}) and a fitness evaluation module. The fitness evaluation includes a HW performance evaluator and a constraint checker (\autoref{fig:system}(a)).

\textbf{HW Performance Evaluator.}
We evaluate all design points using an open-source HW performance evaluator, MAESTRO~\cite{maestro}, which has detailed micro-architectural HW models and is validated against chip prototypes. MAESTRO takes in the accelerator design (HW and mapping) and outputs a detailed HW performance report including latency, area, power, energy, and so on.
In the evaluation, we use HW resource area as the constrained design budget which includes the area of PEs, L1 and L2 buffers\footnote{\label{sec:area_footnote}A complete chip area will also include NoCs, logics, routing, clock trees, pin placement, and others, which need other physical layout considerations and are not considered in MAESTRO and this paper.}. \textbf{Constraint Checker.}
In constraint checker, if the required resources (e.g., area or power) of the proposed design point is larger than the provided budget, we invalidate the design point by re-assigning it a negative fitness value.

\vspace{-2mm}
\subsection{Encoding of Design Point}
\label{sec:encoding}
\subsubsection{\textbf{Encoding: Design Space Description}}
One of our key contributions is a customized encoding, as shown in \autoref{fig:system}(b-c) (notation shown in \autoref{fig:system}(g)), 
to describe an accelerator design-point. 
Our encoding captures the compute resources, mapping, and the memory hierarchy. 
We show a 2-hierarchy level accelerator in \autoref{fig:system}(b-c) as an example. Each key-value pair represents a \textit{gene}. L1-config (yellow) shows the accelerator configuration (HW and mapping) of a 1-D PE array. $\pi_{L1}$, a HW parameter, shows the length of the 1-D PE array. The rest of the genes ($P$,$C$,$K$,$Y$,$X$,$R$, and $S$) describe the mapping parameters for the 1-D PE array, including tiling, order, and parallelism. 
The value genes describe the tile sizes of the corresponding key dimension.
The order of key genes describes the compute order. $P$ gene tells the dimension to parallelize the compute across 1-D PE array.
L2-config shows the HW and mapping across several 1-D PEs arrays, effectively describing a 2-D PE array. $\pi_{L2}$, a HW parameter, shows the number of instantiated 1-D PE arrays, while the rest mapping genes decide the mapping parameters across 1-D PE arrays. Similarly, a 3-level hierarchy (i.e., several 2D arrays) can also be described.

\subsubsection{\textbf{Decoding: Design Point Derivation}}
\label{sec:decoding}
When evaluating the fitness/performance of each individual's genes, we decode them back to an exact accelerator design point. \autoref{fig:system}(d)(e) shows the decoded accelerator of the proposed design point by \autoref{fig:system}(b)(c), respectively. The $\pi_{L2}$ and $\pi_{L1}$ genes decide the PE array sizes and aspect ratio. Therefore different PE arrays are configured in \autoref{fig:system}(d)(e). The $P$ values of L2 and L1 implies how the compute are fetched and parallelized to the PE arrays, e.g., K-C parallelism and X-Y parallelism in \autoref{fig:system}(d)(e). The order of the genes decides the computation order for each tile in the PE arrays. Finally, tile sizes and levels of hierarchies (or called clustering) (e.g., two levels of cluster/ hierarchy in \autoref{fig:system}(b)(c)) determine the \textit{minimum buffer requirement} to house weight, input, and output tensor at both L2 and L1 buffers.


\vspace{-2mm}
\section{Optimization Algorithm}
\label{sec:algorithm}

\subsection{Leveraging Existing Algorithms}
With the thriving of AutoML, 
many optimization algorithms have been developed for automatically searching through a given design-space. They achieve state-of-the-art performance across many domains, including neural architecture search, AI-controlled game, chip design~\cite{google_ai_chipdesign}, and so on. \coopt provides a generic interface enabling us to plug and play many of these existing algorithms (which we leverage from nevergrad~\cite{nevergrad}), as shown in the experiments in \autoref{sec:other_opt}).

However, as discussed in \autoref{sec:challenge}, the design space of HW-Mapping co-optimization is un-smoothed and extreme large. This challenges the search efficiency of the optimization algorithms and makes some of them ineffective under limited sampling budgets (\autoref{sec:other_opt}). This motivates our proposed algorithm, which customizes a genetic algorithm.

\vspace{-2mm}
\subsection{Background of Genetic Algorithm.}
\label{sec:ga_background}
In genetic algorithm (GA), we often call an encoded value of a candidate a \text{gene}, each encoded candidate: an \textit{individual}, a bag of candidates: a \textit{population}, and one iteration of optimization loop: a \textit{generation}. Baseline GA has two standard genetic operators: crossover (blend the genes of individuals and reproduce populations for the next generation) and mutation (perturb the genes of each individual).

Research shows GA reaches competitive performance with deep reinforcement learning \cite{uberGA,openai_es}, and hyper-parameter optimization problem. 
Comparing to many optimizations methods, GA is light, fast, and highly parallelizable~\cite{uberGA, openai_es}. However, the key challenge is its sample efficiency.

GAMMA~\cite{gamma} is an open-source genetic algorithm, tuned for DNN mapping optimization over given HW configurations of accelerators (\autoref{sec:related_work}). 
Despite the effectiveness of GAMMA as a mapping search tool, using it naively to search for HW resources and mapping together will result in a two-loop optimization, which is extremely inefficient, as discussed in \autoref{sec:challenge}. 

\vspace{-2mm}
\subsection{DiGamma: Domain-aware Genetic Algorithm}
\system uses the encoding presented 
in \autoref{sec:encoding} to describe design-points. It then perturbs these 
to search through the co-optimization space.
Rather than using conventional genetic operators (crossover, mutations) to perturb the genes arbitrarily, which is shown to have poor sample efficiency (\autoref{sec:ga_background}), we develop \textit{specialized genetic operators} (i.e., optimization operators) for individual HW and mapping genes to capture the structure of the design space, as described next.

\begin{figure}[t]
\begin{center}

\includegraphics[width=1\linewidth]{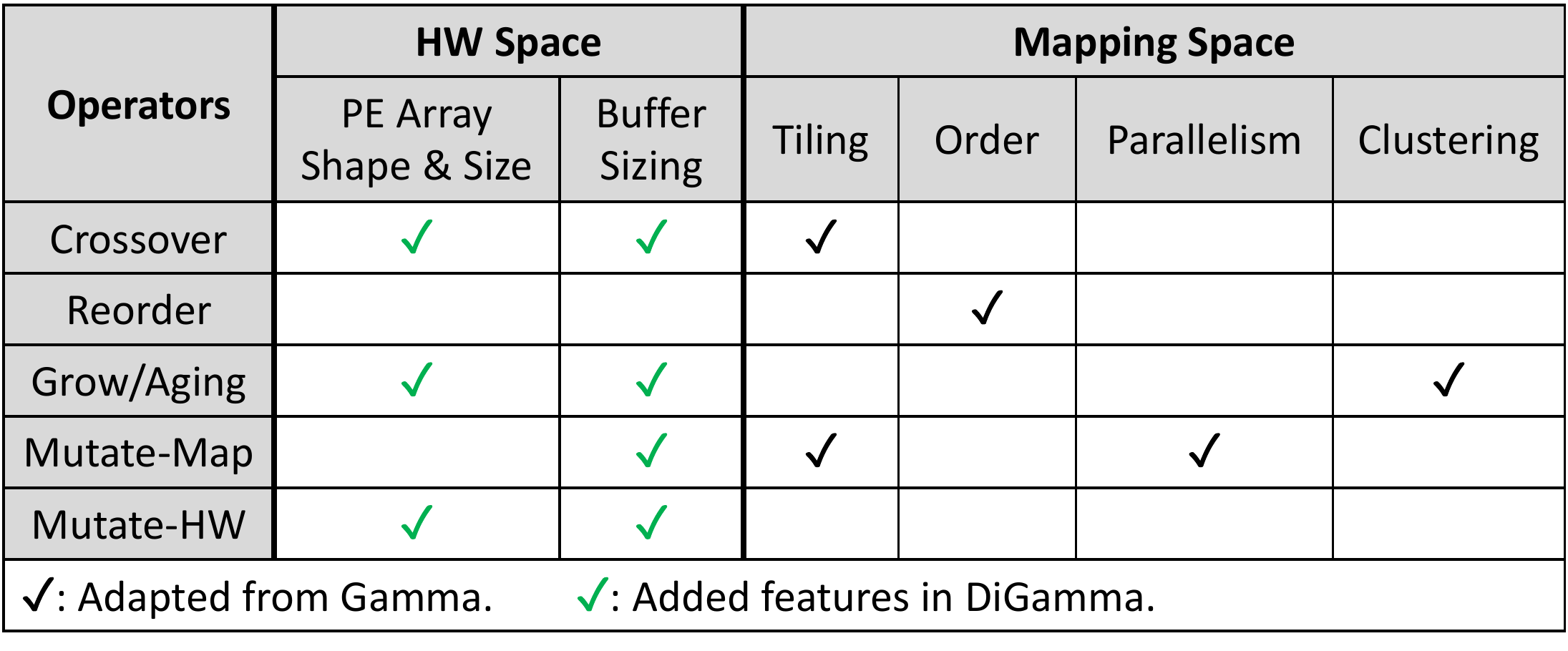}
\end{center}
\vspace{-0.4cm}
\caption{\system's genetic operators\textsuperscript{\dag} and their perturbing space\textsuperscript{\ddag}.}
\footnotesize{\dag: \textbf{Mutate-HW}: Mutation operating on HW space, which tweaks the PE size/shape and also affects the allocated buffer. \textbf{Mutate-Map}: Mutation operating on mapping space and co-affecting buffer choices in HW space.}

\footnotesize{\ddag: Definition of each space can be found in \autoref{fig:related_work_table}.}
\vspace{-0.1cm}
\label{fig:operators}
\end{figure}

The genetic operators responsible for mapping (tiling, order, parallelism, clustering) are modified from GAMMA~\cite{gamma}. Additionally, we implement a \textit{HW} genetic operator to perturb the PE configuration, where the values of $\pi_{L2}$ and $\pi_{L1}$ decide the total number of PEs and the aspect ratio of PE array. Further, for L1 and L2 buffer sizes, we employ a buffer allocation strategy to decide the most optimized buffer allocations of a given individual (\autoref{fig:system}(b)(c)) by the \textit{minimum buffer requirement} derived at decoding block (\autoref{sec:decoding}), i.e., we allocate the exact amount of buffer needed at both L2 and L1 to maximize buffer utilization. We summarize the developed specialized genetic operators and their different perturbing ability across HW and Mapping space in \autoref{fig:operators}.

\begin{figure*}[t]
\begin{center}

\includegraphics[width=1\linewidth]{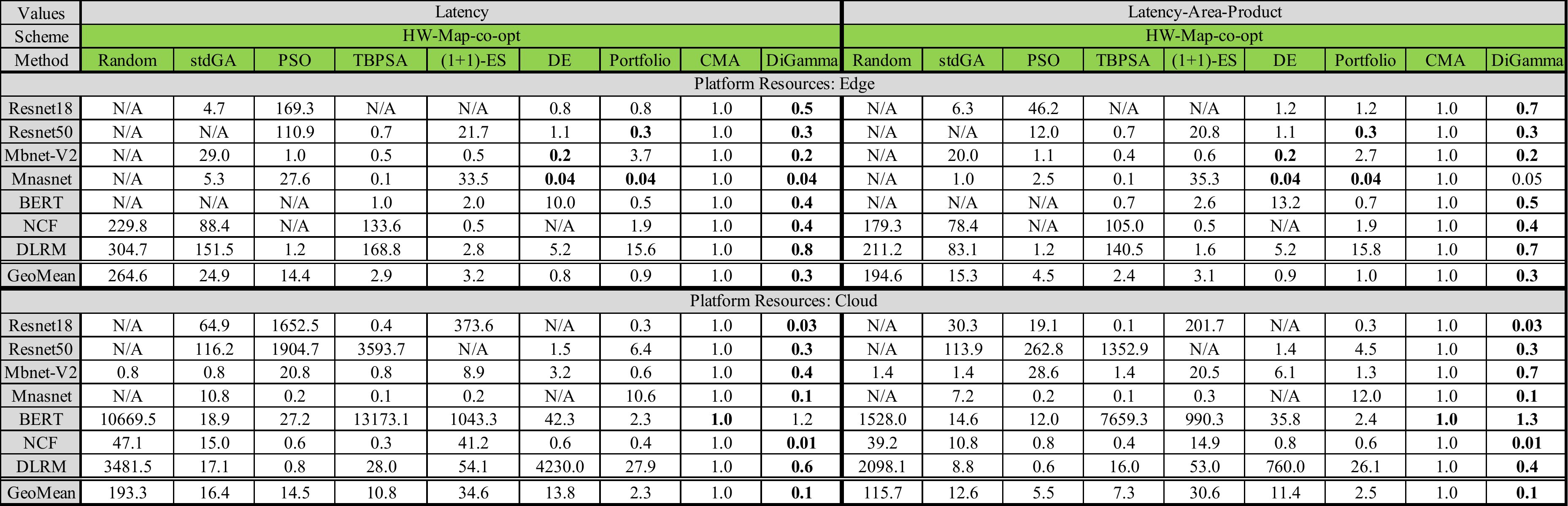}
\end{center}
\vspace{-0.3cm}
\caption{Performance comparisons of different optimization algorithms. Both latency and latency-area-product are normalized by the values of CMA, the best-performing baseline (lower is better). We highlight the best performing algorithm in different tasks (DNN models) in bold. N/A means the algorithm cannot find valid solution that fits in the area constraint under the set 40K sampling budgets.}
\vspace{-0.7cm}
\label{fig:exp_others_latency_LAP}
\end{figure*}

\begin{figure}[t]
\begin{center}

\includegraphics[width=1\linewidth]{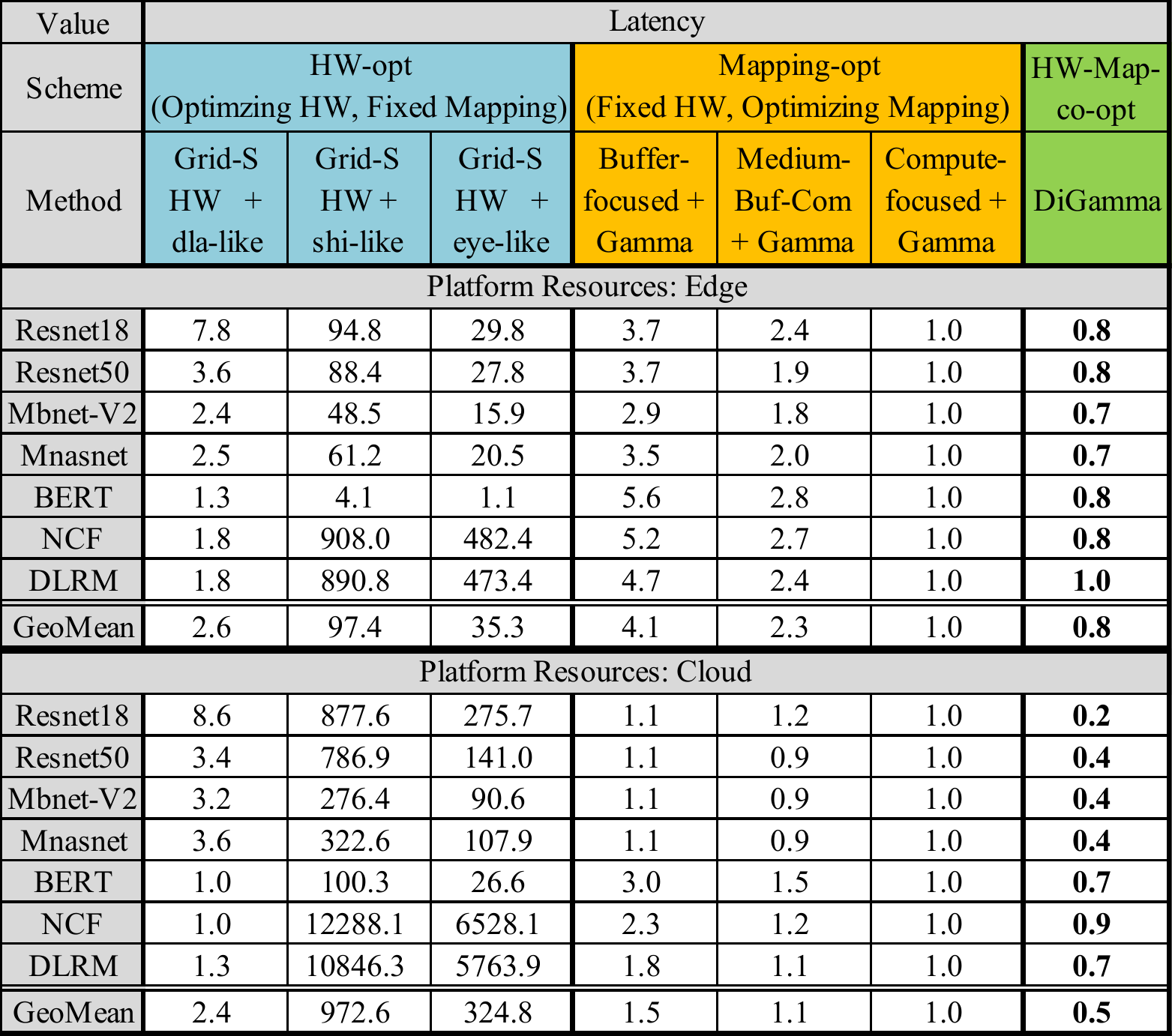}
\end{center}
\vspace{-0.3cm}
\caption{Latency of the found solution by different optimization scheme. Latency values are normalized by the values of best-performing baseline method (Compute-focused + Gamma). Grid-S: grid search. Buffer-focused: large buffer design. Compute-focused: large PE arrays design. Meidum-Buf-Com: medium buffer and PE arrays design.}
\vspace{-0.1cm}
\label{fig:exp_fixed_latency}
\end{figure}

\begin{figure}[t]
\begin{center}

\includegraphics[width=1\linewidth]{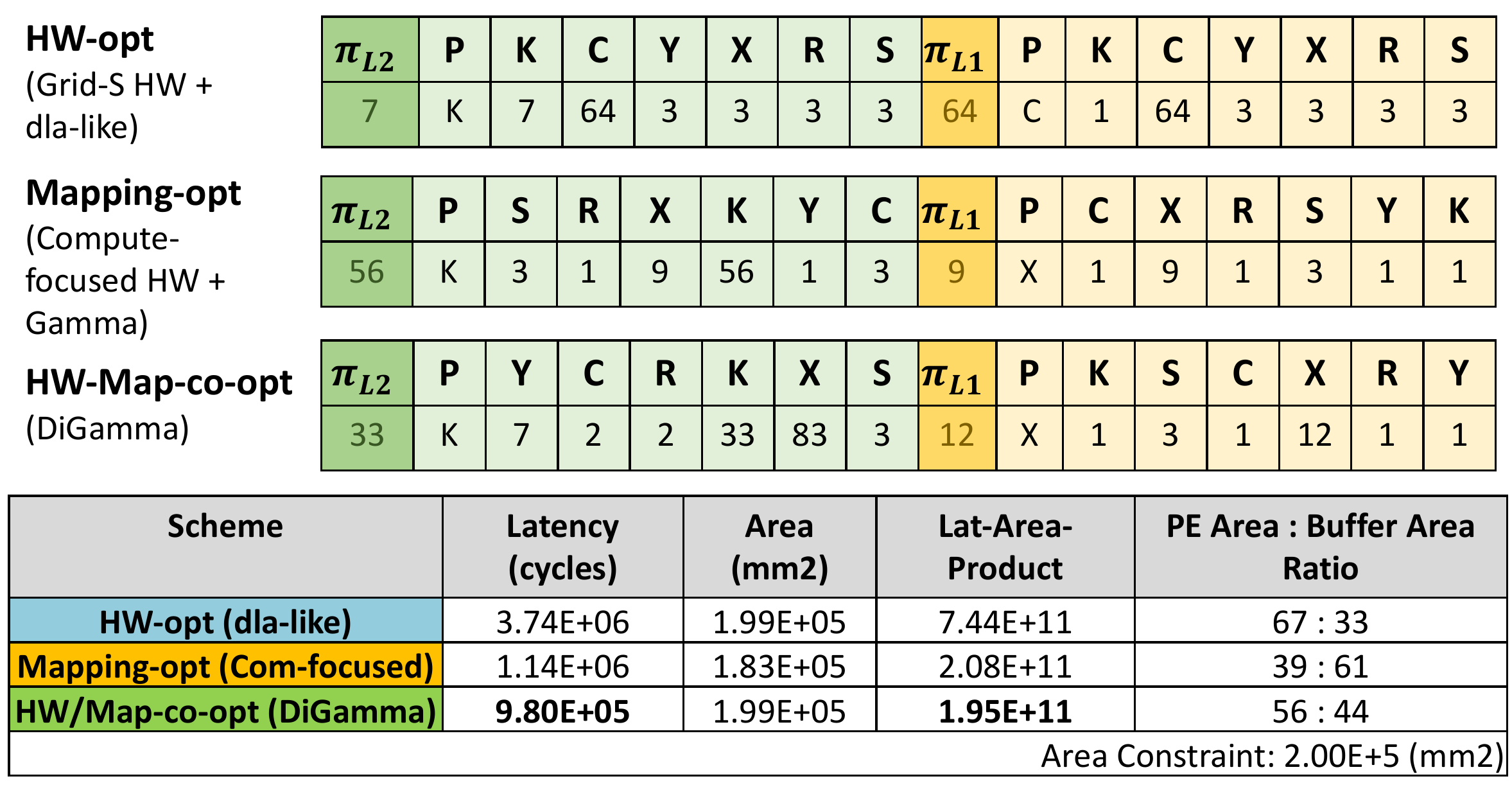}
\end{center}
\vspace{-0.4cm}
\caption{The solution found by different optimization schemes and their corresponding performance on Mnasnet at edge resources.}
\vspace{-0.2cm}
\label{fig:sol_example}
\end{figure}

\vspace{-2mm}
\section{Evaluations}
\label{sec:evaluation}
\vspace{-1mm}

\subsection{Setup}
\label{sec:setup}
Across our experiments, we use the optimizing objective of minimum latency which becomes the performance metric when evaluating the quality of the found solution. Other objectives can also be specified to \system such as power, energy, EDP.

\textbf{DNN Models.}
We experiment 7 DNN models across 3 popular DNN applications: vision (MobilenetV2, Resnet18, Resnet50, Mnasnet), language (BERT), and recommendation (DLRM, NCF).  

\textbf{Edge/ Cloud Platform Resources.}
We evaluate accelerator design under two types of platform resources: edge and cloud. We set the chip area budget for area of PEs and on-chip buffers as \emph{$0.2mm^{2}$ for accelerators in edge~\cite{simba, eyeriss_isca}} and \emph{$7.0mm^{2}$ for accelerators in cloud~\cite{simba}}.

\textbf{Area Cost Model.}
To estimate area cost, we implemented RTL of the various components in \autoref{fig:system}(d-e), synthesized them using Synopsys DC with Nangate 15nm library and 
used Cadence Innovus for place-and-route.
We synthesized the SRAM buffers with SAED32 education library from Synopsys.

\textbf{Sampling Budget of Optimization Algorithms.}
We set the sampling budget (maximum number of sampled points throughout the search process) as 40K points for all the investigated optimization algorithms. For \system, it means population size times number of generations cannot exceed 40K, which takes about 20 mins of CPU-time. 

\textbf{Baseline Optimization Algorithms.}
We take 8 other optimization algorithms, which are widely-used and achieving state-of-the-art performance across different tasks, as baselines. The algorithms include: Random search, standard GA (stdGA), Particle Swarm Optimization (PSO), Test-based Population-Size Adaptation (TBPSA), (1 + 1)-evolution strategy ((1+1)-ES), Differential Evolution (DE), Passive Portfolio (Portfolio), and Covariance matrix adaptation evolution strategy (CMA).

\textbf{Baseline HW and Mapping Optimization Schemes.}
We formulate two kinds of HW and mapping optimization schemes and compare them with \system, listed as follows. 
\squishlist
\item \textbf{HW-opt}: optimizing HW while mapping is fixed. The HW is optimized by grid search approach over number of PEs and buffer sizes. Note that the entire HW configuration design space is as large as $O(10^{12})$, which is hard to enumerate through, and therefore we use grid search. For mapping, we use the manual-designed NVDLA (dla)-like ~\cite{nvdla}, ShiDianNao (shi)-like~\cite{du2015shidiannao}, and Eyeriss (eye)-like~\cite{eyeriss_isca}. 

\item \textbf{Mapping-opt}: optimizing mapping while HW is fixed. The mapping is optimized by GAMMA~\cite{gamma}, a mapping optimizer for a given HW configuration. We cherry-picked three sets of HW configurations: Buffer-focused (small compute + large buffer), compute-focused (large compute + small buffer), and Medium-Buf-Com (medium Buffer + medium compute) configuration for both edge and cloud settings. Note that the design is area constrained, therefore compute and buffer resources are traded-off with each other. 

\item \textbf{HW-Map-co-opt}: using \system to co-optimize both HW and mapping\footnote{The hyper-parameters of \system, (mutation rate, crossover rate, elite ratio, population size to number of generations ratio, and so on), are decided by a Bayesian optimization-based 
search process~\cite{bayes_python}.}.
\squishend

\vspace{-2mm}
\subsection{Comparisons with Baseline Optimization Algorithms}
\label{sec:other_opt}

\autoref{fig:exp_others_latency_LAP} shows the achieved performance (latency) by different optimization algorithms. Note that the proposed \coopt is a supportive back-end and can work well with many state-of-the-art algorithms such as DE, Portfolio, and CMA. 
Considering both stability (without N/A) and performance, CMA is the best-performing one among the compared baseline algorithms. The performance value of CMA represents the best performance that \coopt could bring before discussing the new algorithm, \system. Therefore, we normalize the values in \autoref{fig:exp_others_latency_LAP} by the value of CMA. We walk through more detail of \autoref{fig:exp_others_latency_LAP} and discuss \system's performance, next.

\textbf{At edge settings}, some algorithms can achieve compatible performance with \system in specific tasks such as DE and Portfolio in Mnasnet, however not stable with the respect to the stability across tasks (\autoref{fig:exp_others_latency_LAP}). E.g., DE did not find any solution in NCF, and Portfolio performed 15.6x worse than CMA in DLRM. Besides the latency value of different solutions, we also show their corresponding latency-area-product, since some solution/designs could trade-off areas for better latency. E.g., in BERT cases, TBPSA ranks 2nd across 8 baseline algorithms in latency performance. However, with the respect to latency-area-product, TBPSA is the best among baseline algorithms, meaning it has better area efficiency to achieve similar latency performance comparing to others. Moreover, the poor performance of standard GA contrasts the effectiveness of \system's domain-aware optimization operators. \textbf{At cloud settings}, the wider design space in cloud cases increases the complexity of the optimization tasks. Multiple algorithms are not able to find any valid solutions or can only achieve relatively bad performance. In addition, some performance-leading algorithms such as DE, Portfolio, and CMA at edge settings, become unstable and have much larger swing of achieved performance across different tasks/models at cloud settings. In contrast, \system can stably achieve compatible or better performance than others. Overall, comparing to best-performing baseline algorithm (CMA, Portfolio), \system achieves (geomean) \textbf{3.0x} and \textbf{10.0x} better latency performance at edge and cloud settings, respectively.


\vspace{-2mm}
\subsection{Comparisons with Baseline HW-Mapping Schemes}
In \autoref{fig:exp_fixed_latency} we found that, among the compared methods, using heuristic HW configuration (Compute-focused) with existing mapping searching tool (GAMMA) can yield the best performance, whose value is thus used to normalized the values in \autoref{fig:exp_fixed_latency}. It represents the best achievable relative performance gain before proposing \coopt and \system. We describe more detail of \autoref{fig:exp_fixed_latency}, next.

\textbf{Comparing with HW-opt.} In this experiment, we model the scenario that the researchers designed an optimized mapping for certain DNN models such as NVDLA~\cite{nvdla}, ShiDianNao~\cite{du2015shidiannao}, and Eyeriss~\cite{eyeriss_isca} mapping and want to explore their performance across different models. However, different tasks/ DNN models expose different characteristics (e.g., in general, CNNs are more compute-intensive, and recommendation models are more memory-intensive). To achieve better performance, the researcher could apply optimization algorithms to search for optimal HW configurations (PEs and buffers). Here, we use the grid search approach to search through different PEs and buffers configurations. In \autoref{fig:exp_fixed_latency}, we could observe that \system constantly achieve better performance across three different HW-opt methods, where \system is (geomean) \textbf{3.25x} and \textbf{4.8x} better than the best-performing method (Grid-S + dla-like), in edge and cloud settings, respectively.

\textbf{Comparing with Mapping-opt.} In this experiment, we model the scenario that the researchers designed a mapping optimization algorithm, however relying on a pre-defined HW configuration, which become an inductive bias from the human. For example, different researchers will design different balances between compute and memory resources, where we model with three sets of configurations (\autoref{sec:setup}). Note that the previously effective strategy of exhaustive grid search does not fit this scenario, since grid searching HW plus mapping optimizations will form two loops of the optimization process, whose required sampling budget and search time increase exponentially. In \autoref{fig:exp_fixed_latency}, we can observe that \system is \textbf{1.25x} and \textbf{2.0x} better than the best-performing method (Compute-focused + Gamma), in edge and cloud settings, and more importantly, without the need of human-in-the-loop to cherry-pick the HW configurations.

\vspace{-2mm}
\subsection{Explanation of Found Solutions}
\autoref{fig:sol_example} shows three solutions of different optimization schemes for Mnasnet at edge resources. In HW-opt, we could observe the output/input-channel (K-C) parallelism features of dla-like mapping. In Mapping-opt, we could observe the mapping optimizer find an unique mapping strategy, channel and activation (K-X) parallelism, which is different from dla-like (K-C), shi-like (Y-X), and eye-like (Y-R, row-stationary). \system also finds a mapping with K-X parallelism for Mnasnet, however with better compute and buffer balance, therefore achieving 3.8x and 1.6x better performance than HW-opt and Map-opt.

\vspace{-2mm}
\section{Conclusion and Takeaway}
DNN accelerator design often requires extensive 
tuning for HW resources and mapping due to a large diversity in DNN models and huge design-spaces for HW and mapping.
Recent works show the benefit of either optimizing HW configurations (HW-opt) or mapping configurations (Mapping-opt) independently can harvest several order performance gains compared to fixed HW and mapping. This work shows that co-optimizing both HW and mapping together can (1) yield another tens to hundreds of performance gain comparing to HW-opt and Mapping-opt, (2) more importantly, within the same sampling budget as previous schemes, and (3) largely reduce the repeated engineering cost by minimizing the human decisions in the optimization/design loop of new DNN accelerators for new DNN models, especially, in the era of 
fast-evolving DNNs.

\bibliographystyle{IEEEtranS}
\bibliography{main}

\end{document}